\pdfoutput=1

\documentclass[11pt]{article}

\usepackage[utf8]{inputenc}
\usepackage[preprint]{acl}

\usepackage{times}
\usepackage{latexsym}
\usepackage{amssymb}
\usepackage{booktabs}
\usepackage{tabularx}
\usepackage{xcolor}
\usepackage{colortbl}
\usepackage[T1]{fontenc}

\usepackage[utf8]{inputenc}

\usepackage{microtype}

\usepackage{inconsolata}

\usepackage{graphicx}
\usepackage{booktabs}

\title{\textit{Hire Your Anthropologist!}\\Rethinking Culture Benchmarks Through an Anthropological Lens}

\author{
  Mai AlKhamissi$^{1}$\thanks{\hspace{0.5em}Equal contribution.} \quad
  Yunze Xiao$^{1}$\footnotemark[1] \quad
  Badr AlKhamissi$^2$ \quad
  Mona Diab$^1$ \\[1ex]
  $^1$Carnegie Mellon University \quad
  $^2$EPFL \\
  \texttt{\{malkhami,yunzex,mdiab\}@andrew.cmu.edu} \quad
  \texttt{badr.alkhamissi@epfl.ch}
}

\usepackage[utf8]{inputenc}
\begin{document}
\maketitle
\begin{abstract}

Cultural evaluation of large language models has become increasingly important, yet current benchmarks often reduce culture to static facts or homogeneous values. This view conflicts with anthropological accounts that emphasize culture as dynamic, historically situated, and enacted in practice. In this position paper, we qualitatively examine 20 handpicked cultural benchmarks to show the breadth of how culture is framed within NLP. First, we introduce a four-part framework that categorizes how benchmarks frame culture, such as \textit{knowledge}, \textit{preference}, \textit{performance}, or \textit{bias}. Using this lens and identify six recurring methodological issues: including treating countries as cultures, overlooking within-culture diversity, and relying on oversimplified survey formats. Drawing on established anthropological methods, we propose concrete improvements: incorporating real-world narratives and scenarios, involving cultural communities in design and validation, and evaluating models in context rather than isolation. Our aim is to guide the creation of cultural benchmarks that move beyond simple recall tasks and more faithfully reflect how models respond to complex cultural situations.

\end{abstract}

\section{Introduction}

\begin{figure}[ht]
    \centering
    \includegraphics[width=0.75\linewidth]{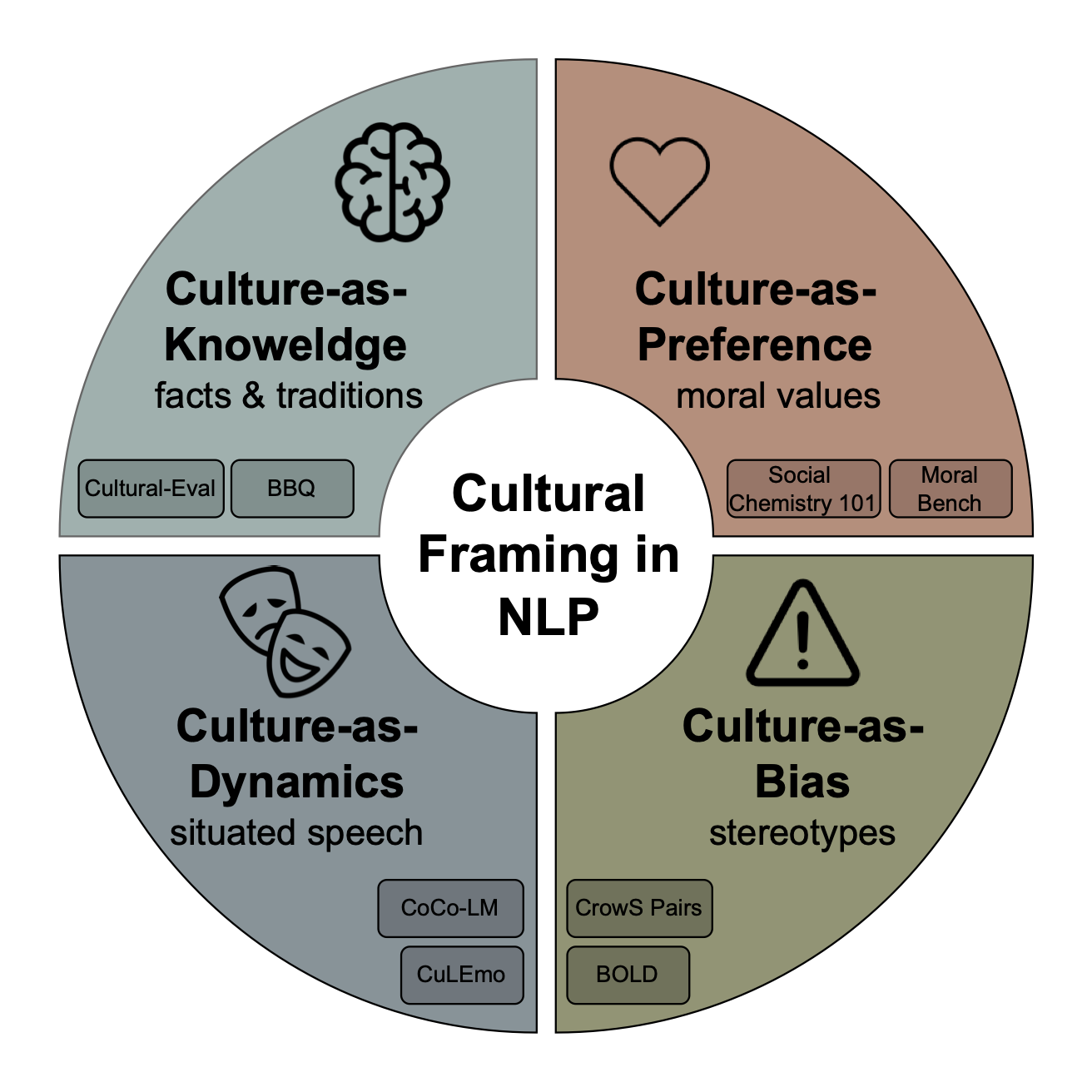}
    \caption{
        \textbf{Cultural Framing in NLP.}
        Our taxonomy of how culture is framed in NLP evaluation. Each quadrant represents a distinct theoretical lens on culture: defining what it entails, illustrating how it is expressed, and providing two representative benchmarks for each framing.
    }
    \label{fig:main}
\end{figure}

Large language models (LLMs) are now being implemented in translation systems \cite{zhu-etal-2024-multilingual}, educational tools \cite{sonkar-etal-2024-pedagogical}, search engines \cite{ziems-etal-2023-large}, and generative platforms that engage diverse public opinions across cultural, linguistic, and political contexts \cite{ziems-etal-2023-large}. As these models become more central to knowledge production and everyday decision-making, it is crucial to assess their sensitivity to cultural nuance\cite{xiao2025humanizingmachinesrethinkingllm}. In response, NLP researchers are constructing benchmarks that aim to measure \textit{cultural competence}: the ability of models to respond appropriately to regionally specific norms, moral frameworks, idioms and socio-political identities.

However, in this growing body of work, the concept of \textit{culture} is often treated as a background variable rather than as a central analytical concern. Benchmark designers rarely engage with anthropology, a discipline that has developed rich tools to understand how culture is lived, contested, and transmitted. Anthropologists study culture not as a fixed set of values or national traits but as a dynamic and situated process shaped by history, power, and everyday meaning-making \cite{Geertz1973}.

To bridge the gap between NLP and anthropology, we analyze 20 cultural benchmarks used in NLP. We argue that while these benchmarks address important questions, such as moral diversity, stereotype reproduction, and regional variation---they often rely on reductive or decontextualized models of culture. 
Consequently, our contribution is as follows: (1) We examine how culture is \textit{conceptualized} within benchmarks, developing a taxonomy of cultural framings used by NLP researchers; (2) We show how six common design choices oversimplify culture, reducing it to surface traits or stereotypes, and restricting which forms of cultural knowledge are recognized or ignored; (3) We offer recommendations for building benchmarks that reflect culture as a lived experience.

\section{Methods}

The choice to analyze 20 benchmarks reflects a methodological commitment to qualitative, humanistic traditions, as opposed to conducting a large-scale survey over many more benchmarks. This (our) approach foregrounds epistemologies, assumptions, and cultural politics often obscured in large-scale corpus analyses. Rather than prioritizing breadth, it emphasizes interpretive depth, reflexivity, and contextual understanding, aligning with scholarship that highlights the limits of scalable methods for capturing nuance, bias, and power dynamics in benchmarks. Our goal is to provide conceptual clarity rather than statistical coverage. Scholars such as \citeauthor{blodgett-etal-2024-human} and voices in ACM interactions \citep{liao-etal-2024-ux-rai,widder-kneese-2025-salvage-anthro-nlp} stress that meaningful cultural analysis demands slow, interdisciplinary, and critically reflexive engagement, a position this study adopts through its carefully curated sample.

Methodologically, many of these datasets are constructed with expert or native annotator input, guaranteeing high-quality, culturally nuanced data that reflect real-world complexities. This set is chosen because it offers a more holistic and representative coverage of cultural dimensions, while addressing current research gaps in cross-cultural NLP, and is aligned with state-of-the-art evaluation needs. They incorporate rigorous annotation protocols, rendering this curated set a more robust and scientifically justified foundation for comprehensive cultural analysis in NLP.

\section{Background and Related Work}

\subsection{What is \textit{Culture}}


Anthropologists have historically studied culture through immersive qualitative methods such as ethnography, long-term fieldwork, and participant observation. Rather than reducing culture to a set of traits or norms, anthropologists emphasize its fluidity, contestation, and entanglement with everyday practices. Thus, the concept of culture has evolved significantly in the social sciences, shifting from fixed categories to fluid and dynamic interpretations. Early anthropological work, notably Boas' cultural relativism \citep{boas1911mind}, rejected universal hierarchies and emphasized understanding each culture on its own terms. Boas argued against simplistic "trait-list" models, proposing instead that culture should be understood historically and contextually.

Subsequent interpretive approaches, epitomized by \citet{Geertz1973}, further reframed culture as "webs of significance" woven by human beings. Geertz emphasized that cultural analysis requires a "thick description," the detailed interpretation of meanings embedded in everyday actions, rituals, and symbols. This perspective shifted the focus from cultural traits toward the interpretive labor necessary to understand situated symbolic practices. The interpretative approach gave way to the "writing culture" turn, where culture is seen as an object that exists outside of the process of writing it. It is not an object to be discovered "out there," but it is constructed through the process of writing about culture \citep{CliffordMarcus1986WritingCulture}. This approach merges researcher with subject, showing that there can be no objective understanding of culture outside the eyes and pen of the researcher. 

Building on these foundations, \citet{goffman1959presentation} introduced the notion of culture as performative and interactional, suggesting that cultural meanings are continuously enacted and negotiated through everyday interactions and social performances. Rather than viewing culture as something individuals inherently possess, this conception emphasizes that culture emerges dynamically in social contexts, shaped by the roles people assume and the interactions they engage in. We use these frameworks to create the taxonomy of culture. 

In this paper, we operationalize these ideas through four evaluation lenses that align with common NLP task designs: what is known (knowledge), what is valued or preferred (preference), how meaning is enacted in context (dynamics), and where identity-linked harms emerge (bias).

\subsection{Culture Benchmarks in NLP}

NLP research has developed benchmarks to evaluate the handling of cultural knowledge using language models, but these works often define \textit{culture} inconsistently and simplistically compared to anthropological scholarship.  Although NLP researchers increasingly recognize that linguistic and social norms vary across cultures and are underrepresented in current resources, significant gaps remain between computational approaches and social science perspectives on culture. The purpose of this work is to foster interdisciplinary dialogue between these fields.

Recent work on cultural and bias evaluation has produced diverse datasets investigating the concept of culture. CultureAtlas \cite{fung2024massivelymulticulturalknowledgeacquisition} broadens multilingual evaluation, integrating everyday cultural knowledge across diverse global contexts, while SeaExam \cite{liu2025seaexamseabenchbenchmarkingllms} emphasizes reasoning and consistency specifically in Southeast Asian scenarios. Furthermore, CDEval \cite{wang-etal-2024-cdeval} systematically maps six cultural dimensions to seven domains, while LLM-GLOBE \cite{karinshak2024llmglobebenchmarkevaluatingcultural} adapts the well-established GLOBE cultural values framework to evaluate open-ended LLM outputs.

Datasets focused on normative and moral reasoning continue to expand, exemplified by Social Chemistry 101 \cite{forbes-etal-2020-social}, which distills common-sense social norms from real-world interactions; and MoralBench   \cite{ji2024moralbenchmoralevaluationllms} provides explicit ethical dilemmas for detailed evaluation. Moreover, NormAd \cite{rao-etal-2025-normad} systematically assesses how models judge the acceptability of everyday etiquette scenarios in 75 countries.

Parallel efforts in bias evaluation are significantly diversified: BBQ \cite{parrish-etal-2022-bbq} identifies stereotypical biases in question-answering tasks; CrowS-Pairs \cite{nangia-etal-2020-crows} targets implicit biases in masked language models; Social Bias Frames \cite{sap-etal-2020-social} annotates offensive implications in language model outputs; and BOLD \cite{BOLD} evaluates bias in open-ended biographical generation. The VL-BiasBench \cite{wang2024vlbiasbenchcomprehensivebenchmarkevaluating} and multilingual M5 \cite{schneider2024m5diversebenchmark} suites extend bias assessment to vision language models, while M-Bias \cite{raza-etal-2024-mbias} offers data-driven interventions to reduce toxicity without sacrificing contextual coherence.

Finally, novel datasets are increasingly expanding into affective and application-specific areas: CuLEmo \cite{belay2025culemoculturallensesemotion} benchmarks emotion recognition across cultures, CASA \cite{qiu2025evaluatingculturalsocialawareness} evaluates social and cultural awareness of LLM-powered web agents, and X-FACTR \cite{jiang-etal-2020-x} assesses multilingual factual knowledge through a cloze-style fill-in-the-blanks question in 23 typologically diverse languages.

\begin{figure*}[ht!]
    \centering
    \includegraphics[width=.8\linewidth]{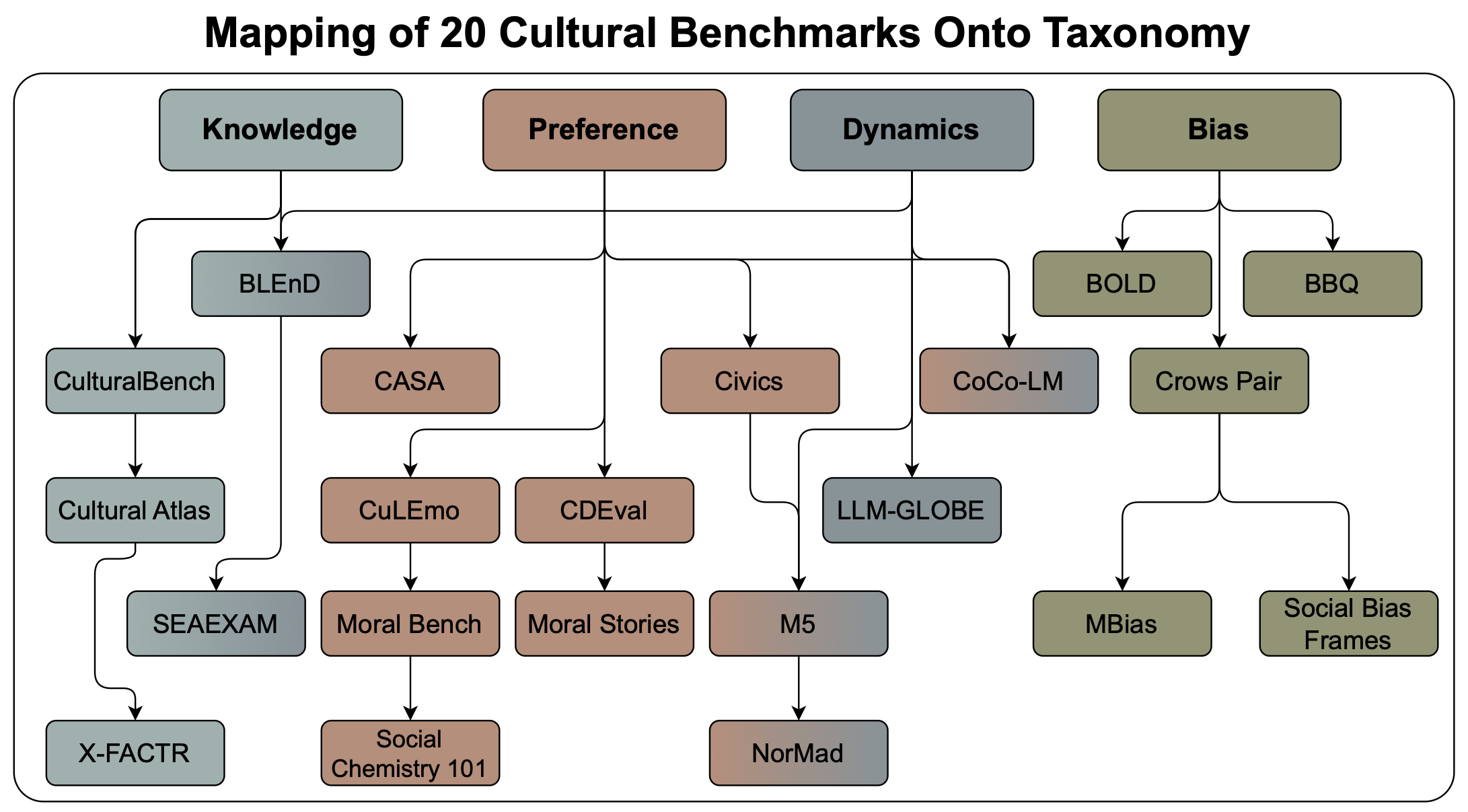}
    \caption{
        \textbf{Mapping of 20 Cultural Benchmarks to the 4 Cultural Dimensions}
        We map the benchmarks analyzed in this work onto the taxonomy proposed in \S\ref{sec:tax}. Each benchmark inherits the dimension of its parent(s).
    }
    \label{fig:taxonomy}
\end{figure*}

\subsection{Critiques within NLP}

Recent work in NLP has exposed fundamental weaknesses in the conceptualization of culture in cultural benchmarks. In the literature, there is a consensus that current evaluation frameworks reduce culture to survey-derived constructs that privilege western-centric assumptions while failing to capture dynamic, situated, and contested cultural meanings \cite{liu-2024-model}. \citet{hershcovich-etal-2022-challenges} identifies unequal cultural representation 
advocate for targeted sampling and Distributionally Robust Optimization to prevent cultural generalization from being disproportionately shaped by dominant groups. \citet{zhou2025culturetriviasocioculturaltheory} extends these concerns by critiquing the over-reliance on nationality-based categorizations and rigid survey methods that flatten dynamic and intersectional cultural identities, proposing more context-sensitive and stereotype-aware datasets to reflect lived cultural realities.

Several influential papers have further articulated these shortcomings. For example,  \citet{chiu2024culturalbenchrobustdiversechallenging} highlight how existing benchmarks are often static, relying heavily on sources such as Wikipedia, and are constrained by narrow survey formats that limit the representation of cultural diversity. These practices risk amplifying urban and western-centric narratives while marginalizing underrepresented perspectives. To counteract this, they introduce CulturalBench, a benchmark designed through human-authored and verified questions that span multiple regions and cultural dimensions, with independent annotator validation to improve robustness and inclusivity. In the same vein, Culture is Not Trivia  \citep{zhou2025culturetriviasocioculturaltheory} critiques the over-reliance on nationality and rigid survey methods, arguing that such approaches flatten dynamic and intersectional cultural identities. They propose more context-sensitive and stereotype-aware datasets to better reflect lived cultural realities and support inclusive benchmark design.


\paragraph{Mitigation Attempts} 

Researchers have attempted to address these limitations through technical interventions, most notably cultural prompting and semantic augmentation. For example, \citet{alkhamissi-etal-2024-investigating} proposed \textit{ anthropological prompting} to make LLMs better reflect the opinions of digitally underrepresented groups. Similarly, \citet{Tao2024} shows that explicitly instructing LLMs to adopt specific cultural perspectives can significantly reduce Western bias in most countries and territories, although this approach sometimes introduces new biases and rests on the problematic assumption that culture can be cleaned up by demographic targeting. Although these innovations show promise in controlled settings, they fundamentally maintain the flawed premise that culture can be distilled through survey data or prompt engineering, an assumption increasingly questioned within the research community.

\subsection{Existing Frameworks for Cultural Evaluation}

The framework presented by \citet{vijay-hershcovich-2024-abstract} provides an important foundation for cross-cultural NLP by mapping out four main cultural elements: linguistic form, common ground, aboutness, and values, to guide researchers through the conceptual complexities of working with cultural diversity. Their approach highlights strategic challenges and high-level pathways for ensuring more fair and representative NLP systems. In contrast, our work expands this foundation by focusing directly on how culture is operationalized in evaluation metrics and benchmark design.

Concurrent with our work, Adilazuarda et al. (2024) \citeauthor{adilazuarda-etal-2024-towards} offer a high-level taxonomy that distinguishes between demographic and semantic proxies of culture, revealing the field’s reliance on operational shortcuts rather than explicit theorizing  cultural processes. Complementarily, Liu et al. (2023) \cite{Liu+2024} propose a fine-grained taxonomy that organizes cultural elements into ideational, linguistic, and social dimensions, grounded in anthropological definitions and emphasizing the interplay between language and social practice. 

Havaldar et al. (2025) and Pawar (2025) push cultural evaluation beyond static text benchmarks. Havaldar et al. introduce a sociocultural framework for the evaluation of conversational LLM, focusing on \textit{linguistic style} as a key site where cultural differences manifest, and proposing new metrics such as \textit{conversational structure, stylistic sensitivity,} and \textit{subjective correctness} to assess how models adapt across cultural contexts. Pawar (2025) similarly broadens the field by offering a cross-modal taxonomy, connecting textual, visual, and multimodal forms of cultural understanding through dimensions of \textit{representation, knowledge,} and \textit{social sensibility}

Together, these frameworks have provided important structure for an emerging interdisciplinary space, but they also highlight the limits of static categorization. Building on this work, the taxonomy advanced here engages both perspectives: acknowledging the functional distinctions emphasized by Adilazuarda et al. (2024) and the theoretical breadth introduced by Liu et al. (2023). We also draw on more recent developments such as Havaldar et al. (2025), who extend cultural analysis into conversational and stylistic metrics by designing a new benchmark, and Pawar (2025), which integrates multimodal dimensions of cultural understanding. Our framework differs from these by centering the conceptual modeling of culture itself rather than its downstream manifestations in text, interaction, or modality. In our conception, culture is not treated as a set of characteristics, but as a shifting interpretive process situated in discourse, interaction, and power, enabling a richer analysis of how LLMs reproduce and transform cultural meaning across contexts.

\section{Culture Benchmark Taxonomy}
\label{sec:tax}
While technical mitigation strategies have advanced the capture of cultural complexity in NLP, our contribution is to systematically examine how culture itself is conceptualized within leading benchmarks. We begin by offering a taxonomy that clarifies the diverse definitions of culture used across these resources, helping reveal the range of cultural perspectives that inform benchmark design. Building on this taxonomy, we present recommendations in \S\ref{sec:recommendations} to inform future benchmark development and encourage a more deliberate engagement with cultural nuance.
We first introduce a four-part taxonomy for cultural evaluation in NLP: culture-as-knowledge, culture-as-preference, culture-as-dynamics, and culture-as-bias. The taxonomy is both descriptive and prescriptive. Descriptively, it distills how current benchmarks operationalize culture; prescriptively, it guides future work to combine lenses rather than treat them in isolation, encouraging multidimensional designs that better capture cultural nuance.

\paragraph{Culture-as-Knowledge.} Culture consists of facts, traditions, and symbolic references that can be recalled or identified. This lens treats culture as a body of factual or object-level information tied to national or regional identities, for example, benchmarks that assess whether models recognize culturally specific foods, holidays, or practices. This view echoes early \textit{trait list} approaches in anthropology \citep{boas1911mind}. 

\paragraph{Culture-as-Preference.} This angle conceptualizes culture as a set of shared moral, social, or political attitudes, often derived from surveys or aggregate annotations. It reflects a more psychological or normative model of culture as shared values or aggregated moral intuitions, typically measured through survey instruments. Although widely used in cross-cultural psychology \cite{hofstede1980culture,inglehart2005modernization}, this framing has been critiqued in anthropology for reducing culture to a “thin” notion of coherence and boundedness within groups, obscuring internal contestation, hybridity, and change \cite{ortner2006anthropology,AbuLughod1991}. Nevertheless, it remains attractive because it is highly operationalizable: surveys and annotation schemes allow for scalable comparisons across populations and across models, even if this comes at the cost of nuance. 

\paragraph{Culture-as-Dynamics.} Culture is not treated as static, but as an entity that acts in specific contexts through language, narrative, and interaction. It draws on linguistic and symbolic anthropology to understand culture as enacted, improvised, and contextual, shaped by speaker positionality, audience, and occasion \citep{goffman1959presentation,hymes1974foundations,bauman1990poetics}. This view understands culture as something that people actively \textit{ perform} in context, rather than an attribute they possess. It focuses on how meaning is shaped through interaction, making it better suited for capturing how models respond to different social situations. Benchmarks such as CoCo-LM, BLEND, Moral Stories, and CuLEmo exemplify this view by modeling how culture is performed in dialogue, story, or emotion expression.

\paragraph{Culture-as-Bias.} This dimension particularly frames cultures through a series of stereotypes. The goal of this identification is harm reduction and the elimination of bias.  It tests how the model outputs reflect and reinforce stereotypes. We introduce this category specifically for NLP evaluations: culture is inferred from patterns of stereotyping or discriminatory output, especially along dimensions of race, gender, or class. Although these benchmarks play an essential role in fairness research, they often rely on binary identity comparisons without interrogating the cultural production of these categories. We aim to introduce an intersectional lens through this category \cite{Crenshaw1991,alim2009translocal}. BBQ, Social Bias Frames, BOLD, M-Bias, and VL-Bias Bench all operate within this paradigm, focusing on identity-based harm and fairness violations.

\subsection{Contribution}

While \citet{hershcovich-etal-2022-challenges} advocate for broad methodological reflection and community engagement, our taxonomy critically dissects actual evaluation practices, exposing how specific methodological choices shape, limit, or obscure cultural complexity. We provide prescriptive strategies for recording ambiguity, contestation, and dissent as core data signals, and for institutionalizing plural ground truths within benchmarks. Therefore, while their framework maps the strategic landscape of the field, our work intervenes more directly at the level of metrics, offering concrete recommendations for embracing the messiness and multiplicity of lived culture in NLP evaluation. 

In contrast, we interrogate how culture is framed in the evaluation metrics themselves, making visible methodological practices such as annotation, bias assessment, and consensus building that can flatten cultural complexity. By analyzing how culture is already encoded in evaluation and urging benchmarks to record zones of dispute and ambiguity, our taxonomy moves beyond cataloging toward a design roadmap for resources that reflect multiplicity, contestation, and lived experience.
.

\subsection{Mapping Benchmarks Onto the Taxonomy}

We review 20 prominent cultural benchmarks in NLP and map them to the four cultural dimensions. 
Datasets such as \textit{CultureAtlas} \cite{feng-2024-towards}, \textit{BLEnD} \cite{myung2025blendbenchmarkllmseveryday}, and \textit{CulturalBench} \cite{chiu2024culturalbenchrobustdiversechallenging} exemplify \textbf{Culture-as-Knowledge}, focusing on factual recall tied to national or regional identity. \textbf{Culture-as-Preference} is illustrated by benchmarks such as \textit{Social Chemistry 101} \cite{forbes-etal-2020-social}, \textit{MoralBench} \cite{ji2024moralbenchmoralevaluationllms}, and \textit{NormAd} \cite{rao-etal-2025-normad}, where models are evaluated for alignment with aggregated moral or social judgments. \textbf{Culture-as-Dynamics} appears in datasets such as \textit{CoCo-LM} \cite{COCO}, \textit{BLEnD}, and \textit{MoralStories}, foregrounding the context-sensitive nature of language and emotion. Finally, \textbf{Culture-as-Bias} is captured by benchmarks such as \textit{BBQ} \cite{parrish-etal-2022-bbq}, \textit{Social Bias Frames} \cite{sap-etal-2020-social}, and \textit{BOLD} \cite{BOLD}, which examine how models reproduce identity-based harms. 

Figure \ref{fig:taxonomy} presents the category breakdown.
Each of the four parts of our framework highlights a distinct facet of cultural life and can fit different evaluation goals. However, most of the prominent NLP benchmarks to date lean on two narrow frameworks: culture-as-knowledge and culture-as-preference, while treating culture as a set of static facts or simple surveyable attitudes. In contrast, anthropological work views culture as dynamic, situated, and relational. In our recommendations (\S\ref{sec:recommendations}), we will expand on how to use this taxonomy in future work.

\section{Methodological Limitations and Ways Forward} 
\label{sec:limitation}

This section identifies six methodological limitations that flatten cultural complexity in current NLP benchmarks. We argue that because of these methodological choices, the concept of culture is often unidimensional and flattened. 

The limitations are: (a) Platform bias favoring Western, urban demographics; (b) Conflating nation states with cultural boundaries; (c) Treating individual annotators as cultural representatives; (d) Reducing moral reasoning to survey scales; (e) Assuming cultural consensus where disagreement exists; and (f) Stripping context from cultural scenarios. 

These practices systematically undermine all four dimensions of our taxonomy: knowledge, preference, dynamics, and bias, by imposing essentialist frameworks that contradict complex understandings of culture as contested, situated, and fluid. We also show where benchmarks do a good job in finding creative ways around these methodological limitations. Table \ref{tab:overview} maps these limitations across benchmarks, revealing how methodological choices constrain conceptual possibilities and pointing to the need for more reflexive, context-sensitive, and theoretically grounded evaluation methods.

\begin{table*}[ht!]
    \centering
    \renewcommand{\arraystretch}{1.1}
    \setlength{\tabcolsep}{3pt}
    \small
    \begin{tabular}{@{}l|cccccc@{}}
    \toprule
    \textbf{Benchmark} & \textbf{Platform Bias} & \textbf{Nation-State} & \textbf{Annotation} & \textbf{Moral Simplification} & \textbf{Consensus} & \textbf{Lack of Context} \\
    \midrule
    BBQ & \checkmark & -- & \checkmark & -- & -- & -- \\
    BLEND & \checkmark & \checkmark & \checkmark & -- & -- & -- \\
    BOLD & \checkmark & -- & \checkmark & -- & -- & -- \\
    CASA & \checkmark & -- & \checkmark & \checkmark & \checkmark & -- \\
    CDEval & \checkmark & \checkmark & \checkmark & \checkmark & \checkmark & -- \\
    Civics & \checkmark & -- & \checkmark & \checkmark & -- & \checkmark \\
    CoCo-LM & \checkmark & -- & -- & -- & -- & -- \\
    Crows-Pairs & \checkmark & -- & \checkmark & -- & -- & -- \\
    CuLEmo & -- & -- & \checkmark & -- & -- & -- \\
    CulturalAtlas & \checkmark & \checkmark & -- & -- & -- & \checkmark \\
    CulturalBench & \checkmark & -- & \checkmark & -- & -- & -- \\
    LLM-GLOBE & \checkmark & \checkmark & -- & -- & -- & -- \\
    M-Bias & \checkmark & -- & \checkmark & -- & -- & -- \\
    M5 & \checkmark & -- & -- & -- & -- & -- \\
    Moral Bench & \checkmark & -- & \checkmark & \checkmark & \checkmark & \checkmark \\
    Moral Stories & \checkmark & -- & \checkmark & \checkmark & \checkmark & \checkmark \\
    NormAd & \checkmark & -- & \checkmark & \checkmark & \checkmark & -- \\
    SeaExam & \checkmark & \checkmark & \checkmark & -- & -- & -- \\
    Social Bias Frames & \checkmark & -- & \checkmark & -- & -- & -- \\
    Social Chemistry 101 & \checkmark & -- & \checkmark & \checkmark & \checkmark & -- \\
    X-FACTR & \checkmark & \checkmark & -- & -- & -- & -- \\
    \bottomrule
    \end{tabular}
    \caption{\textbf{Presence of Methodological Limitations Across 20 Cultural Benchmarks.} Each column indicates the presence (\checkmark) or absence (--) of a limitation defined in Section~\ref{sec:limitation}: \textbf{Platform Bias} (\S\ref{sec:platform}), \textbf{Nation-State} (\S\ref{sec:nation-state}), \textbf{Annotation} (\S\ref{representation}), \textbf{Moral Simplification} (\S\ref{sec:survey}), \textbf{Consensus} (\S\ref{assume}), \textbf{Lack of Context} (\S\ref{sec:context}).}
    \label{tab:overview}
\end{table*}

\subsection{Platform Bias and Demographic Skew}
\label{sec:platform}

Many cultural benchmarks are drawn from Reddit, X, or Wikipedia, which mostly mirror Western, male, urban, and digitally literate users. This privileges certain demographics over others. Social Chemistry 101 inherits Reddit’s community biases, and datasets like BBQ, BOLD, CulturalAtlas, and others import similar skews; the effect extends to CASA, CDEval, CulturalBench, LLM-GLOBE, SeaExam, and X-FACTR. Treating such homogeneous pools as globally representative risks erasing offline, indigenous, rural, or otherwise marginalized worldviews.

Digital anthropology demonstrates that digitally mediated expressions, such as posts and opinions on social networks, do not automatically reflect the full complexity of 'real social facts', as recognized in ethnographic research \cite{miller2018digitalanthropology}. Miller stresses that digital platforms produce distinctive forms of sociality and cultural interaction, deeply shaped by the technological frameworks and demographics of their users; Interpreting these online traces as direct representations of offline life risks missing the broader social, cultural, and material contexts in which meaning is formed \cite{horstmiller2012digitalanthropology,miller2018digitalanthropology}. Anthropological analysis insists on holistic ethnography, situating digital worlds within the complexities of larger social relations, and cautioning against reducing the mediation of online experience to simple, universal truths \cite{horstmiller2012digitalanthropology,miller2018digitalanthropology}. Digital is not a transparent lens but an active mediator: sometimes intensifying difference, sometimes shaping whose voices become visible or invisible, and always requiring careful contextual understanding to avoid misrepresenting the realities of diverse communities \cite{horstmiller2012digitalanthropology,miller2018digitalanthropology}.

Partial remedies exist. FLEAD (Federated Learning to Exploit Annotator Disagreements) \citep{rodriguez-barroso-etal-2024-federated} models each annotator’s perspective to surface minority viewpoints that majority aggregation can suppress, yet its reach is limited by the sparse annotator metadata and the exclusion of communities without digital access. 

\subsection{Nation-State as a Proxy for Culture}
\label{sec:nation-state}

Several NLP benchmarks adopt nationality as the primary cultural framework, assuming that country boundaries are mapable to homogeneous cultures. SeaExam and CulturalAtlas assign norms by country with statements such as 'In Chinese culture...', while BLEND, CDEval, LLM-GLOBE and X-FACTR rely on national groups that obscure internal ethnic, linguistic and religious differences. SeaExam aggregates Singapore, Malaysia, and Thailand under one label, and LLM-GLOBE conflates language with culture (for example, Arabic as “Middle Eastern culture”), overlooking diasporic contexts. In agreement with our work,\citeauthor{havaldar2025cac} also shows how, instead of defining cultures, LLMs instead use proxies using nation states.

Social science scholarship underscores why such assumptions are fundamentally problematic. \citeauthor{gupta1992beyond} argues that nation states are politically constructed entities, shaped by transnational flows, colonization histories, and shifting political economies, rather than natural or coherent cultural units. Treating national borders as cultural boundaries flattens complex identities and histories, ignoring the dynamic, overlapping, and contested nature of culture that ethnographic research reveals. This critique aligns with broader calls within anthropology and cultural studies to move beyond the nation state as the default unit of cultural analysis and instead to address the multiplicity and fluidity of cultural formations \cite{gupta1992beyond,appadurai1996modernity}.

BLEND demonstrates progress by combining intra- and cross-national perspectives, and Jiraibench \citep{xiao2025jiraibenchbilingualbenchmarkevaluating} investigates how meanings change between discourse types, platforms, and diasporic communities. Jiraibench is an exploratory first step; sustained research is needed to develop richer, more situated evaluations that reflect contested, plural, and evolving cultural life.

\subsection{Representation Fallacy Through Annotation}
\label{representation}

Many benchmarks assume that a few annotators can represent the cultural norms of a group. This affects BBQ, BLEND, CASA, CDEval, SEAExam, LLM-GLOBE and others. SEAExam collapses Singapore, Malaysia, and Thailand into one label, and LLM-GLOBE rarely reports annotator demographics. Treating such labels as comprehensive removes the variation by class, gender, generation, and religion. Using LLMs as “cultural judges” without external validation amplifies the problem.

As \citet{abu-lughod1991writing} and \citet{mahmood2005politics} note, representation is not only who speaks, but also how voices are situated within power relations. Little is understood of how annotation choices, sampling practices, or positionality of annotators shape the contours of what counts as 'culture' in these data sets. This methodological opacity risks reinforcing normative assumptions, perpetuating a narrow view of cultural diversity that can marginalize less visible voices and mask tensions within communities. Without reflexive attention to which experiences are foregrounded and which are silenced, benchmarks may inadvertently produce a shallow or exclusionary portrait of cultural life.

\subsection{Moral Simplification through Survey Format Overload}
\label{sec:survey}

Many “cultural reasoning” benchmarks reduce Culture-as-Preference and Culture-as-Dynamics to survey responses, collapsing nuanced judgments into Likert scales or binaries. CASA, CDEval, Civics, Moral Bench, Moral Stories, NormAd, and Social Chemistry 101 rely on simplified prompts that flatten context. Moral Bench, grounded in Moral Foundations Theory built from WEIRD samples, treats ethics as consensus scores and projects a narrow frame as universal. Anthropological work shows why this fails: moral life is context-dependent, narratively situated, and historically contingent, not a set of fixed rules \citep{zigon2007moral, laidlaw2013subject}.

Hybrid approaches such as Social Bias Frames \cite{sap-etal-2020-social} replace surveys with scenario-based tasks that present culturally contingent dilemmas. These reveal value conflicts flattened in traditional formats, though scenarios often reflect Western paradigms of autonomy rather than collectivist frameworks \citep{hofstede1980culture}. Progress requires preferring open-ended tasks that solicit narrative explanations alongside ratings.

\subsection{Assumption of Cultural Consensus}
\label{assume}

Many benchmarks treat aggregated judgments as correct answers, assuming cultural agreement on norms, which eliminates the contested nature essential to Culture-as-Dynamics. CASA, CDEval, Moral Bench, Moral Stories, NormAd, and Social Chemistry 101 use majority votes as ground truth. Social Chemistry 101 collects explanations showing diverse reasoning, but still trains models on the most common answers. This treats disagreement as noise, rather than recognizing culture as a site of ongoing negotiation. As \citet{gal2015politics} and \citet{tsing2005friction} emphasize, disagreement is a fundamental feature of cultural life, not a flaw.

FLEAD's federated architecture rejects consensus assumptions by maintaining annotator-specific models and surfacing disagreement \cite{rodriguez-barroso-etal-2024-federated}. This reveals cultural fractures, such as divergent interpretations of offenses between religious groups. However, broad demographic labels still mask intragroup diversity \cite{blodgett-etal-2020-language}. Recent work by \citet{dai2025embracingcontradictiontheoreticalinconsistency} and \citet{fleisig2024majoritywrongmodelingannotator} points toward disagreement-aware protocols that recognize contradiction as essential; we applaud this line of work and recommend building on it. 

More promising directions treat culture as situated and temporal, emphasizing multiplicity over universality. BLEND \cite{myung2025blendbenchmarkllmseveryday} advances this by evaluating culturally situated knowledge in 16 regions and 13 languages using contemporary community-informed data, and SEACrowd \cite{lovenia-etal-2024-seacrowd} shows what large-scale community collaboration can achieve with more than 400 Southeast Asian contributors.

\subsection{Decontextualized Prompts and Abstracted Norms}
\label{sec:context}

Cultural benchmarks often test models on questions disconnected from historical and political context, which strips away the situational grounding necessary for Culture-as-Dynamics. For example, BBQ presents stereotype questions without a grounding in social histories, CASA assumes universal norms, and CulturalAtlas reduces culture to decontextualized trivia \cite{zhou-2024-towards}. By isolating questions from their environments, these benchmarks treat cultural reasoning as context-free, overlooking how norms are formed and contested within specific conditions.

Some benchmarks move toward scenario-based tasks that embed challenges in locally meaningful contexts \cite{bravansky2025rethinkingaiculturalalignment}. However, these often rely on generic scenarios that lack microcontextual cues such as tone, history, or power dynamics \cite{blodgett-etal-2020-language}. The most promising direction involves observing model behavior rather than just output. Following \citet{li2025actionsspeaklouderwords}, who found that models with minimal stated bias still exhibit \textit{significant sociodemographic disparities} in simulations, the evaluation must recognize the gaps between the stated results and contextual decisions.

\citet{blodgett-etal-2024-human} argue the abstraction of culture in benchmark design preserves the illusion of neutrality, but erases contestation and dissent, privileging dominant perspectives and rendering marginal ones invisible. The most promising anthropological approach moves evaluation toward observing model behaviors within context-rich simulations and attending to gaps between stated outputs and embedded decisions. Recent work demonstrates that models can exhibit significant disparities along sociodemographic lines in complex settings, even when explicit bias is minimal. 

\section{Recommendations}
\label{sec:recommendations}

These six design choices arise from the drive for scalability and quantifiability of NLP, which conflicts with the anthropological notions of situated practice. Here is our list of recommendations for ways forward. 

\subsection{Social Science Collaboration}

Our central recommendation is to establish ongoing, methodical collaborations that position social scientists as integral partners throughout every stage of AI and NLP research. Drawing on blueprints developed in interdisciplinary initiatives and early prototypes from our own work, we propose a pipeline that spans problem definition, data collection, benchmark design, annotation, and evaluation, with social scientists actively co-leading each phase. Our ongoing project seeks to formalize such practices, providing concrete steps and governance models for co-designing research agendas, annotation guidelines, and evaluation protocols to ensure that social scientific expertise is embedded and reflected in every methodological decision.

An example of this approach is \citeauthor{alkhamissi-etal-2024-investigating}, who developed interdisciplinary annotation protocols and evaluation frameworks by establishing sustained collaboration between computer scientists and social scientists from the inception of the project through analysis, illustrating both challenges and best practices to integrate anthropological insights into the design of the NLP benchmark. This work, along with surveys such as \textit{NLP for Social Good} \citep{karamolegkou2025nlpsocialgoodsurvey}, \textit{Anthropology and AI: A Framework for Mutually Beneficial Collaboration} \citep{artz2025anthropologyAI}, and other initiatives, underscore that a deeply embedded social science partnership leads to richer, more robust and culturally attuned AI systems.

\subsection{Participatory Design}

Cultural benchmarks must evaluate models in dynamic and interactive settings that resemble ethnographic observation. BLEND progresses through the generation of community-informed scenarios, while action-based approaches assess the behavior of the model in simulated environments. The field should deploy ethnographic testing, capture model adaptation strategies and norm negotiation, embed micro-contextual cues, and co-design scenarios with cultural insiders. Rather than static Q\&A, benchmarks should trace how models navigate competing norms and make implicit judgments in open-ended interactions.

\subsection{Decoupling Culture from Nation}

Decoupling culture from the nation-state within AI systems requires moving beyond the assumption that national borders neatly align with cultural boundaries and instead recognizing the internal diversity, hybridity, and mobility present within and across nations. Although the category of the nation can remain analytically useful, benchmarks and models should treat it as one layer among many, situating national identity alongside transnational, diasporic, and local cultural affiliations. This can be achieved by designing evaluations that explicitly compare how core concepts (such as freedom, kinship, or authority) are deployed across legal, religious, activist, and everyday discourses, and by collaborating with local and transnational communities to identify relevant categories, axes of variation, and migration histories. Through such pluralistic methods, AI systems can avoid essentializing culture as nation-bound while leveraging the nation category for context-specific inferences, regulatory needs, or critical comparisons where appropriate.

\subsection{Critical Annotation}

A critical annotation approach must move beyond simply measuring annotator disagreement, recognizing that annotation itself is a social and interpretive act deeply shaped by the power relations and positionality within annotation work. While existing tools for calculating inter-annotator agreement help quantify variation, this is only a partial solution: a single annotator's choices reflect not the entirety of a culture or religion, but their own relationship with and perspective on those traditions---a performance influenced by social, economic, and institutional contexts. Drawing from anthropology's tradition of 'writing culture' \citep{CliffordMarcus1986WritingCulture}, critical annotation recommends protocols that allow ambiguity and narrative, incorporating continuous scales, narrative storytelling, and space for dissent and self-reflection rather than forcing binary consensus. Methods should encourage iterative, community-driven processes where annotation guidelines and ethical dilemmas are refined collaboratively, capturing multiple, contested meanings and the complexity of lived experience. 

\subsection{Mapping Controversy}
Tools should be developed to systematically map zones of controversy in datasets, institutionalizing plural 'ground truths' where contestation and disagreement are recognized as valuable data signals. Rather than erasing dissent, benchmarks should record, analyze, and represent sites of active dispute as key sources of interpretive richness. Empowering communities to identify which disagreements matter and leveraging critical annotation practices shifts evaluation away from artificial consensus and allows AI systems to better reflect the multiplicity and contested nature of culture. 

Examples of controversy mapping and plural ground truth approaches can be found in interdisciplinary algorithm studies and recent NLP benchmarks that explicitly document disagreement. \citet{munk2024beyond} have mapped algorithmic controversies in the scientific literature by tracking how different communities contest and reinterpret algorithmic outputs, making dissent part of the data landscape. In NLP, FLEAD's consensus-building protocol preserves annotator differences, and projects like Social Chemistry 101 retain gradient judgments to reflect multiple interpretations of social norms rather than flattening to a single answer.

To improve on these methods, future benchmarks can systematize the documentation of annotation rationales, support longitudinal annotation so that evolving viewpoints are tracked, and provide transparent metadata about power structures and institutional roles in the annotation process. Involving community members directly in the identification and contextualization of disputes will further root ground truths in lived experience.

\subsection{Using the Full Taxonomy }

Our taxonomy serves as an integrative framework that bridges persistent disconnects between the diverse conceptualizations of culture found in NLP research, evaluation design, and anthropological theory. By systematically distinguishing and relating culture-as-knowledge, culture-as-preference, culture-as-dynamics, and culture-as-bias, it reveals both methodological fragmentation and the potential for meaningful synthesis in future benchmarks. We recommend using the full taxonomy of culture when thinking about how to approach the question of culture within NLP. We also encourage researchers not only to use all four dimensions of this taxonomy when conceptualizing culture in NLP, but also to place it in conversation with other major frameworks: \citeauthor{adilazuarda-etal-2024-towards} on demographic and semantic proxies; \citeauthor{liu-etal-2025-culturally} on ideational, linguistic and social dimensions; and \citeauthor{pawar2024survey} on multimodal and interactional approaches to cultural awareness. When used together, these frameworks can advance both theoretical and evaluative sophistication, linking conceptual precision with methodological design. Moving forward, benchmark development should integrate all aspects of our taxonomy while engaging these complementary models to produce more dynamic, intersectional, and contextually grounded evaluations of how NLP systems represent and negotiate culture.

\section{Conclusion}
Current NLP benchmarks often reduce culture to trivia, to fixes such as diverse annotators, localized fine-tuning, and disagreement-aware metrics capture only fragments. Drawing on anthropology, we propose a flexible taxonomy and an interdisciplinary pipeline where social scientists collaborate with NLP researchers across design, data, modeling, and evaluation. Social theory can guide scenario-based, context-rich prompts and treat annotator disagreement as a signal, while NLP methods scale these insights. Sustained collaboration replaces ad hoc work with iterative co-design and joint interpretation, yielding benchmarks that reflect fluid identities and models that engage culture as lived practice, improving both bias mitigation and interpretability.

By bridging critical social science with NLP practice, our framework not only critiques the status quo, but also offers a roadmap for more robust, inclusive, and dynamic cultural evaluation. It calls for models that do not simply conform to survey-driven metrics of alignment, but meaningfully engage with cultural difference as lived, negotiated, and plural, co-produced with the communities they are meant to serve. 

\section*{Limitations}

Our study provides a comprehensive qualitative critique of existing cultural benchmarks and offers concrete desiderata for building more realistically grounded cultural benchmarks based on deep anthropological tenets. However, we acknowledge some critical limitations. A significant constraint lies in the scope of the benchmarks analyzed. Although we reviewed 20 cultural benchmarks, our selection was necessarily limited to publicly available datasets and widely recognized frameworks. As a result, other relevant benchmarks, particularly those that are emerging or proprietary, may provide additional insights that were not covered in this work. In particular, we did not review all benchmarks that work to address these issues. Several key datasets, such as the PRISM alignment dataset \cite{kirk2025prism}, Value Kaleidoscope \cite{sorensen2024value}, and Social Norms in Cinema \cite{rai2024social}, have made significant strides toward overcoming anglocentric biases and embracing pluralistic cultural perspectives. However, we excluded them from our analyses to maintain a bird's eye view focused on categorizing broader types of critique in benchmarks, rather than conducting an exhaustive dataset inventory.

Our study focuses primarily on benchmarking methodologies rather than assessing the real-world impact of cultural benchmarks on LLM applications. Although improving benchmarks is a crucial step towards cultural sensitivity in AI, future research should investigate how these advancements translate to tangible improvements in AI-human interactions across various domains, including education, governance, healthcare and cross-cultural communication. Without real-world deployment analysis, the long-term effectiveness of improved cultural benchmarks remains an open question.

In addition, our position paper focuses on text-based evaluations, omitting vision-oriented culture benchmarks that interrogate the understanding of models of imagery, symbolism, and geospecific visual practices. Multimodal datasets such as CulturalVQA (visual question answering on food, clothing and rituals drawn from 11 countries) \cite{nayak-etal-2024-benchmarking}, CVQA \cite{romero2024cvqaculturallydiversemultilingualvisual} (a culturally diverse and multilingual VQA set spanning 28 nations and 26 languages), and WorldCuisine \cite{winata-etal-2025-worldcuisines} (a dataset of world cuisines) demand analytical framework (e.g. visual-semantics grounding, region-specific iconography, and image–text alignment biases) that differ fundamentally from the linguistic constructs underlying our framework. Incorporating them would therefore require distinct taxonomies and fairness diagnostics, which we flag as an important avenue for future multimodal culture work. 

We also recognize that we are not the first to introduce a taxonomy of cultural understanding in NLP \cite{liu-etal-2025-culturally}. However, in contrast, our four-part taxonomy distinguishes between culture-as-knowledge (factual/symbolic information), culture-as-preference (shared values and moral attitudes), culture-as-dynamics (performative, enacted, context-dependent practices), and culture-as-bias (patterns of stereotyping and harm reduction). Our framework is not only descriptive, mapping how culture is conceptualized, but also prescriptive, challenging single-axis approaches, and recommending that future benchmarks intentionally combine these dimensions to better capture the complexity and contested nature of cultural life. 

We recognize the value of concurrent work toward shared definitions, but claim that our four-part taxonomy and critical evaluation analysis provide a foundation for both understanding and transforming cultural benchmarking, moving from descriptive surveys to actionable recommendations that place contestation and plural ground truths at the heart of NLP evaluation.

Despite these limitations, we believe that this work serves as an important step toward rethinking cultural benchmarking in AI. 

\section*{Ethical Statement}

This position paper advocates for a rethinking of cultural benchmarks in Large Language Models (LLMs) through an anthropological lens, a perspective that introduces several ethical considerations regarding the framing, methodology, and potential impact of our analysis.

The issue of intellectual ownership and participatory ethics is relevant to this discussion. By advocating for the incorporation of ethnographic data and lived experiences into AI benchmarks, this paper indirectly raises concerns about how cultural knowledge is sourced and used. Ethical AI research must avoid extractive approaches that commodify cultural knowledge without consent or compensation. The recommendations presented in our paper should be interpreted as a call for ethically co-designed benchmarks rather than an endorsement of appropriating cultural insights without the involvement of the communities they pertain to.

Furthermore, this paper discusses bias and misrepresentation in AI, but operates within the constraints of currently available benchmarks. The selection of 20 cultural benchmarks, while intended to be comprehensive, is still shaped by the availability of research, the predominance of sources in English, and our own methodological choices. Any analysis of AI benchmarks carries the risk of reinforcing certain academic and institutional perspectives while overlooking others. Future interdisciplinary research should seek to include voices from non-Western epistemologies, indigenous knowledge systems, and community-driven AI governance models.

In general, this paper seeks to approach cultural benchmarking in AI with ethical responsibility, advocating methodologies that are rigorous and respectful of the diverse human experiences they seek to represent. However, the very act of critiquing, selecting, and proposing cultural benchmarks must itself remain subject to ethical scrutiny, ensuring that the pursuit of cultural sensitivity in AI does not inadvertently replicate the biases and epistemic imbalances it aims to address.

\bibliography{latex/custom,anthology_0,anthology_1}

\clearpage

\appendix

\section*{Appendix}

\section{Case Study: CASA vs BLEND}


This section contrasts the BLEND and CASA benchmarks, highlighting their distinct strengths in evaluating cultural understanding in large language models (LLMs).

BLEND excels in assessing \textbf{culture-as-knowledge}. It immerses models in realistic conversational scenarios that demand the contextual recall and appropriate application of culturally specific facts. For example, prompts might involve characters discussing preparations for Lunar New Year or Diwali, testing the model's ability to refer to customary foods, greetings, or rituals in natural dialogue. This tests the integration of factual knowledge, such as understanding that joss paper is burned in Chinese ancestral rituals or that modaks are made during Ganesh Chaturthi.

In contrast, CASA is particularly strong in capturing \textbf{culture-as-preference}. It presents models with survey-style statements reflecting social or moral attitudes and assesses whether the models align their responses with the presumed dominant values of particular cultural groups. For example, CASA tasks might ask, "In your culture, is it acceptable to speak loudly in public spaces?" or assess whether a model expresses greater agreement with statements endorsing filial piety in East Asian contexts versus valuing individual autonomy in Western contexts.

However, both benchmarks operationalize culture in a certain way. BLEND’s interactive prompts assess cultural expertise by integrating object-level details that a model must integrate into natural discourse, reflecting authentic practice and symbolic reference. For example, BLEND might present a scenario in which characters plan a traditional celebration, such as discussing foods for the feast of Chuseok or customs for Ramadan, challenging the model to mention appropriate dishes (like exchanging songpyeon in Korea or breaking fast with dates in Muslim contexts). Meanwhile, CASA uses responses to attitude questions (such as the acceptability of public displays of emotion) to compare value patterns across cultural domains by mapping responses to expected normative judgments (like stoicism in Japanese settings versus expressiveness in Mediterranean cultures). Although BLEND challenges the breadth of LLM by introducing multiple details to recall for a culture, CASA may oversimplify by overlooking contestations or differences between cultures.

\section{Description of Reviewed Benchmarks}
\label{sec:review}

Table \ref{tab:cultural_frameworks} shows the benchmarks reviewed in this work classified using the taxonomy introduced in \S\ref{sec:review}.

\begin{table*}[t]
\centering
\small
\rowcolors{3}{white}{gray!6}
\newcolumntype{L}{>{\raggedright\arraybackslash}X}
\begin{tabularx}{\textwidth}{@{}l L l@{}}
\toprule
\textbf{Benchmark} & \textbf{Description} & \textbf{Citation}\\
\midrule
BBQ & Tests bias robustness in QA across nine U.S. demographic dimensions with ambiguous/disambiguated contexts. &  \citet{parrish-etal-2022-bbq}\\
BLEND & 52.6K culturally grounded Q\&A pairs spanning 16 regions and 13 languages for evaluating everyday knowledge. &  \citet{myung2025blendbenchmarkllmseveryday}\\
BOLD & Benchmarks social biases in open-ended generation across five domains with 23,67=9 prompts. &  \citet{BOLD}\\
CASA & Corpus of culturally aware situational actions testing context-appropriate behavioral selection across societies. &  \citet{liu-etal-2024-casa}\\
CDEval & Cross-domain suite benchmarking cultural adaptability across news, social media, and literature with expert-verified items. &  \citet{wang-etal-2024-cdeval}\\
Civics & Corpus of civic-knowledge questions (history, government, social studies) probing democratic processes and civics literacy. &  \citet{pistilli2024civicsbuildingdatasetexamining}\\
CoCo-LM & Pretraining objective that corrects/contrasts corrupted sequences, improving efficiency on GLUE, SQuAD, and culturally sensitive tasks. &  \citet{COCO}\\
Crows-Pairs & Minimal-pair corpus measuring stereotype direction and strength across social categories. &  \citet{nangia-etal-2020-crows}\\
CuLEmo & Cross-cultural emotion-understanding benchmark across six languages requiring nuanced cultural reasoning. &  \citet{belay2025culemoculturallensesemotion}\\
CultureAtlas & A fine-grained multicultural dataset (127K assertions, 10K+ cities, 2.5K+ ethnolinguistic groups) that enables cultural-knowledge reasoning and norm-violation detection. &  \citet{feng-2024-towards}\\
CulturalBench & 1,227 human-verified questions across 45 regions and 17 topics for evaluating cultural knowledge and adaptability. &  \citet{chiu2024culturalbenchrobustdiversechallenging}\\
LLM-GLOBE & Assesses cultural values in LLM outputs via the GLOBE framework, comparing alignment with societal value systems. &  \citet{karinshak2024llmglobebenchmarkevaluatingcultural}\\
M-Bias & Multilingual corpus probing gender, profession, and cultural stereotypes in 50+ languages. &  \citet{raza-etal-2024-mbias}\\
M5 & Multimodal vision–language benchmark spanning 41 languages to highlight performance disparities. &  \citet{schneider-sitaram-2024-m5}\\
Moral Bench & Evaluates LLMs’ moral reasoning via Moral Foundations Theory, measuring ethical alignment. &  \citet{ji2024moralbenchmoralevaluationllms}\\
Moral Stories & Branching-narrative dataset for grounded social reasoning over norms, intents, and consequences. &  \citet{emelin2020moralstoriessituatedreasoning}\\
NormAd & Measures model adaptability to varying cultural norms through social-acceptability judgments. &  \citet{rao-etal-2025-normad}\\
SeaExam & A benchmark derived from Southeast Asian educational exams for assessing LLM proficiency in regional languages and cultural contexts. &  \citet{liu-etal-2025-seaexam}\\
Social Bias Frames & 150K annotations capturing pragmatic bias implications in language for nuanced social-media detection. &  \citet{sap-etal-2020-social}\\
Social Chemistry 101 & 292K commonsense social norms across 12 dimensions for everyday moral-norm reasoning. &  \citet{forbes-etal-2020-social}\\
X-FACTR & Multilingual fact-checking benchmark with claims in 25 languages to test factual verification and cross-lingual generalization. &  \citet{jiang-etal-2020-x}\\
\bottomrule7e34
\end{tabularx}
\caption{Descriptions and citations for 20 cultural NLP benchmarks, ordered alphabetically by benchmark name.}
\label{tab:cultural_frameworks}
\end{table*}

\end{document}